\documentclass[runningheads]{llncs}
\usepackage[T1]{fontenc}
\usepackage{hyperref}
\usepackage{color, colortbl}
\usepackage{xcolor}
\usepackage{arydshln}
\usepackage{pifont}
\usepackage[symbol]{footmisc}
\usepackage{tablefootnote}
\def\x{{\mathbf x}}

\definecolor{LightCyan}{rgb}{0.88,1,1}

\newcommand{\cmark}{\ding{51}}
\newcommand{\xmark}{\ding{55}}

\usepackage{boldline}
\usepackage{booktabs}
\usepackage{graphicx}
\usepackage{diagbox}
\usepackage{makecell}
\usepackage{multirow}
\usepackage{lipsum}  
\usepackage{physics}
\usepackage{floatrow}

\usepackage{times}
\usepackage{amssymb}
\usepackage{subcaption}
\usepackage{mathtools}
\usepackage{bm}
\usepackage[accsupp]{axessibility}
\usepackage{adjustbox}
\usepackage{cite}
\usepackage{mathtools}
\DeclarePairedDelimiter\ceil{\lceil}{\rceil}

\makeatletter
\def\adl@drawiv#1#2#3{%
	\hskip.5\tabcolsep
	\xleaders#3{#2.5\@tempdimb #1{1}#2.5\@tempdimb}%
	#2\z@ plus1fil minus1fil\relax
	\hskip.5\tabcolsep}
\newcommand{\cdashlinelr}[1]{%
	\noalign{\vskip\aboverulesep
		\global\let\@dashdrawstore\adl@draw
		\global\let\adl@draw\adl@drawiv}
	\cdashline{#1}
	\noalign{\global\let\adl@draw\@dashdrawstore
		\vskip\belowrulesep}}
\makeatother

\newcommand{\mycomment}[1]{}
\begin{document}
	%
	
	\title{Temporal Divide-and-Conquer Anomaly Actions Localization in Semi-Supervised Videos with Hierarchical Transformer}
	\titlerunning{Temporal Divide-and-Conquer Anomaly Actions Localization}
	%
	\author{Nada Osman\inst{1} \and
		Marwan Torki\inst{2}}
	\authorrunning{N. Osman, M. Torki}
	%
	\institute{ 
		University of Padova, Italy\\ \email{nadasalahmahmoud.osman@phd.unipd.it} \vspace{0.3cm} \and Alexandria University, Egypt\\
		\email{mtorki@alexu.edu.eg}\\
	}
	\maketitle              
	\begin{abstract}
		Anomaly action detection and localization play an essential role in security and advanced surveillance systems. However, due to the tremendous amount of surveillance videos, most of the available data for the task is unlabeled or semi-labeled with the video class known, but the location of the anomaly event is unknown. In this work, we target anomaly localization in semi-supervised videos. While the mainstream direction in addressing this task is focused on segment-level multi-instance learning and the generation of pseudo labels, we aim to explore a promising yet unfulfilled direction to solve the problem by learning the temporal relations within videos in order to locate anomaly events. To this end, we propose a hierarchical transformer model designed to evaluate the significance of observed actions in anomalous videos with a divide-and-conquer strategy along the temporal axis. Our approach segments a parent video hierarchically into multiple temporal children instances and measures the influence of the children nodes in classifying the abnormality of the parent video. Evaluating our model on two well-known anomaly detection datasets, UCF-crime and ShanghaiTech, proves its ability to interpret the observed actions within videos and localize the anomalous ones. Our proposed approach outperforms previous works relying on segment-level multiple-instance learning approaches while reaching a promising performance compared to the more recent pseudo-labeling-based approaches.
		
		
		\keywords{Anomaly Actions Localization \and Anomaly Detection  \and Weakly Supervised Learning \and Semi-Supervised Learning \and Hierarchical Modeling \and Transformers.}
	\end{abstract}
	\section{Introduction}
	
	Surveillance represents the backbone of almost all security systems; however, extracting important events, particularly anomalies, from the vast pool of collected videos is a time-consuming and exhaustive task. There arises a critical need for an intelligent system capable of accurately and autonomously extracting and localizing events of interest. The main challenge in training such a localization model lies in the lack of supervision, as the massive amount of collected data for this task is unsupervised or weakly supervised. Consequently, the employed model must be able to decode event sequences, extract embedded relations, and identify potential outlier behaviors.
	Many prior works have tackled this task, some in a fully unsupervised approach~\cite{uns_anomaly1, uns_anomaly2}, but mostly in a semi-supervised approach ~\cite{dataset_anomaly, ss_anomaly1, ss_anomaly2, ss_anomaly3, ss_anomaly4}. In the semi-supervised setting, each video is labeled as normal or anomalous, yet the specific location of the anomaly segment within the video is unspecified. The standard protocol, in this case, operates at the segment level, aiming to maximize the hidden representation gap between normal and anomalous segments \cite{dataset_anomaly, ss_anomaly1, ss_anomaly2}. To further improve the learning process in weakly supervised settings, the research community has started to give more attention to the generation and refinement of per-segment pseudo labels \cite{ss_anomaly3, ss_anomaly4, ss_anomaly5, li2022self, zhang2023exploiting, al2024coarse}. Alternatively, recent attention has been directed towards image reconstruction approaches, wherein models are trained to reconstruct normal videos or frames~\cite{rec_anomaly1, rec_anomaly2, yan2023feature}. Subsequently, these trained models are repurposed to distinguish between high-quality reconstructions, indicative of normal frames, and poor-quality reconstructions, indicative of anomaly frames. These anomaly detection techniques are segment-level or frame-level approaches, often overlooking per-video classification. However, it is reasonable that decoding the relative dependencies within videos should allow for event understanding and anomaly localization. This approach would offer several advantages: 1) Reformulating the anomaly detection task from segment-based to video-based is more compatible with real-world surveillance applications, enabling the processing of longer videos rather than just short segments. 2) It enhances explainability and human understanding of the anomaly event and its temporal context. 3) Compared to pseudo-labeling, it provides a faster end-to-end training process. 4) Hypothetically, it should offer higher generalization ability than pseudo-labeling techniques, which are tailored to specific datasets and may inherit dataset specifics and noise. Therefore, in this work, we explore this direction, proposing a novel approach to utilize per-video classification for anomaly localization.\\
	
\begin{figure}[t]
		\centering
		\includegraphics[width=\linewidth]{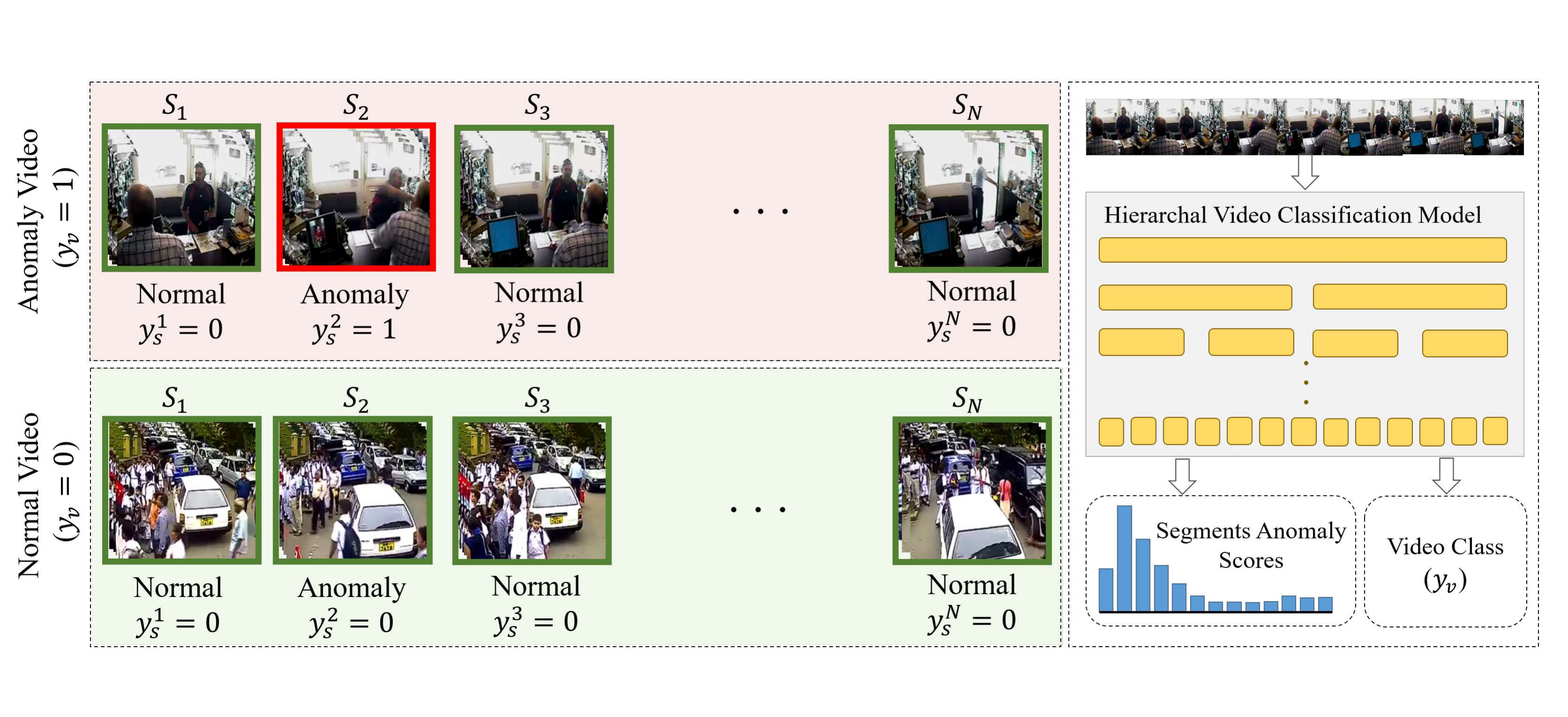}
		\caption{Each video is split into $N$ segments. A normal video ($y_v=0$) contains only normal segments ($y_s^i=0, \forall i\in[1:N]$). While an anomaly video ($y_v=1$) contains at least one anomaly segment ($y_s^i=1, \exists i\in[1:N]$). Our approach employs a hierarchical transformer model to classify the abnormality of the whole video, in addition to producing abnormality scores for the individual segments. This approach differs from previous works that overlook the context of the entire video and classify individual segments independently.}
		\label{fig:problem_statement}
	\end{figure}

	\noindent
	If a model can distinguish videos containing anomalies from normal videos, the model's hidden representation should inherently contain sufficient information about the anomaly locations. We propose a temporal divide-and-conquer transformer-based model to classify the normality of a parent video and its children segments in a hierarchical approach. We extract potential anomaly segments based on the aggregated classifications of the model throughout the hierarchical levels, integrated with the corresponding activation maps. As shown in Figure \ref{fig:problem_statement}, a video is split into $N$ temporal segments, where a normal video does not contain any anomaly events, while an abnormal video must contain at least one anomalous action. Unlike previous works, our objective is to solve two tasks: the per-video classification task, predicting $y_v$, and the per-segment classification, predicting $y_s^i$ for each segment $i$ in the video. The per-segment classification is fully unsupervised, based on the information extracted during the per-video classification. As shown in Figure \ref{fig:problem_statement}, our model processes the input video in a hierarchical manner, where each level of the hierarchy aims to measure the abnormality of the included video patches, reaching up to the final level representing the fine-graned segments. Therefore, the proposed model produces two outputs: a high-level per-video classification and per-segment abnormality scores. Our evaluation results provide promising insights into the effectiveness of our proposed approach in understanding the observed video, extracting the temporal relations, and identifying the anomaly events. The main contributions of our work can be summarized as follows:
	
	\begin{enumerate}
		\item We revise the segment-wise anomaly detection task, transforming it into anomaly localization within videos, allowing the learning procedure to benefit from per-video classification in detecting the anomaly segments in a semi-supervised manner.
		\item We propose our temporal divide-and-conquer transformer-based model that aims to weigh the abnormality of various temporal patches of the video hierarchically in order to provide a fine-grained aggregated estimation of the abnormality of the detected events.
		\item We evaluate our model on two well-known datasets for anomaly detection and conduct different ablation experiments. 
	\end{enumerate}
	
	\section{Related Work}
	
	\subsection{Anomaly Detection}
	Anomaly detection is an important task that has been always gaining attention in computer vision. The task takes three primary learning formats: Fully-supervised learning, where ground truth labels are available for both normal and anomaly actions \cite{rec_anomaly1, s_anomaly1}; Semi-supervised learning, having ground truth labels for entire videos but lacks annotations for anomaly segments within those videos \cite{dataset_anomaly, ss_anomaly1, ss_anomaly2, ss_anomaly3, ss_anomaly4, ss_anomaly5, rec_anomaly2}; and Fully-unsupervised learning, which operates without any ground truth labeling \cite{uns_anomaly1, uns_anomaly2}. While full supervision offers a direct path to problem-solving, the tremendous process of annotating large datasets makes it impractical. Conversely, unsupervised learning bypasses the need for data annotations but introduces heightened complexity to the learning process. Consequently, significant attention has shifted towards the semi-supervised learning format, where models can leverage video annotations to enhance the recognition of anomaly segments within those videos.\\
	
	\noindent
	In the semi-supervised setting, previous works primarily applied multiple instance learning (MIL), where the model pairs one normal video with another anomaly video and trains the model to maximize the representation gap between the two segments. The leading work in this category is \cite{dataset_anomaly}, which aimed to maximize the distinction between the highest-scoring segments within the positive and negative bags, thereby maximizing the inter-bag distance. Building upon this concept, \cite{rec_anomaly2} introduced a novel approach by concurrently minimizing intra-bag distance while maximizing inter-bag distance. Additionally, in \cite{rec_anomaly1}, MIL is once again employed, this time augmented with an attention mechanism aimed at ranking segments within each bag, with the objective of accentuating the disparity between the highest-ranking segments across both bags.\\
	
	\noindent
	Another leading technique for the semi-supervised learning approach is pseudo-labeling. This approach involves generating pseudo-labels for unlabeled segments and subsequently training the model in a fully-supervised manner using these generated labels. In \cite{ss_anomaly5}, a method is proposed where an action classifier module is trained for each segment. This module takes segment images and the video's label as inputs, iteratively refining and purifying the predicted segment classifications. Conversely, in \cite{ss_anomaly4, ss_anomaly3, tian2021weakly} \cite{ss_anomaly3}, multiple instance learning (MIL) is employed to generate pseudo-labels, which are utilized to train the classifier model. With the rising attention to pseudo-labeling approaches, some recent works are dedicated to the generation, refinement, and cleaning process of pseudo-labels, as in \cite{zhang2023exploiting, al2024coarse}. In contrast to these works, we reformulate the problem into anomaly localization within videos, proposing a model that can interpret the temporal axis along input videos, extracting anomaly events.
	
	\subsection{Class Activation Maps Learning}
	
	Class activation maps (CAM) learning is a popular technique in computer vision. It is mainly used in image-based object detection, where CAM-based techniques are widely used in scoring objects within images for object detection and localization \cite{grad-cam, grad-cam++, score-cam, layercam}. Recently, the application of CAM techniques to videos started to emerge. For example, in \cite{tcam}, temporal max pooling is proposed to aggregate per-frame CAMs for video object localization. Another interesting application of CAM learning in videos is \cite{fcam}, where optical flow-based CAM is utilized for weakly-supervised segmentation. In the domain of anomaly detection, \cite{cam_anomly} proposed the utilization of CAM to generate a form of pseudo labeling, integrated into a second learning phase with MIL to create two groups of training segments: positive and negative. The final classification in this approach employs K-nearest neighbor to the positive and negative groups. Similarly, in \cite{tian2021weakly}, segment activations, represented by features magnitude, are utilized to generate top-k normal/anomaly segments that are then used to train a per-segment classifier for anomaly detection. Our objective in this work is to generate different types of activation maps, which are then fused together to yield an attentive estimation of the abnormality associated with the events observed within the classified video.
	
	\subsection{Temporal Hierarchical Modeling}
	Understanding and extracting temporal dynamics from videos and real-world time series pose significant challenges. Consequently, techniques like temporal multi-scale and hierarchical modeling are important in processing temporal data. For instance, in \cite{slowfast}, the proposed model operates across two temporal scales, slow and fast, resulting in more robust representation extraction. Numerous studies have adopted hierarchical modeling methods for real-world temporal data, such as \cite{th1}, or for video tasks, such as action recognition \cite{th2, th3}. Therefore, in our work, we utilize hierarchical modeling along the temporal axis to leverage temporal information extracted from input videos at varying temporal resolutions using a transformer-based model.
	
	\section{Temporal Divide-and-Conquer Approach}
	
	\begin{figure}[t]
		\centering
		\includegraphics[width=\linewidth]{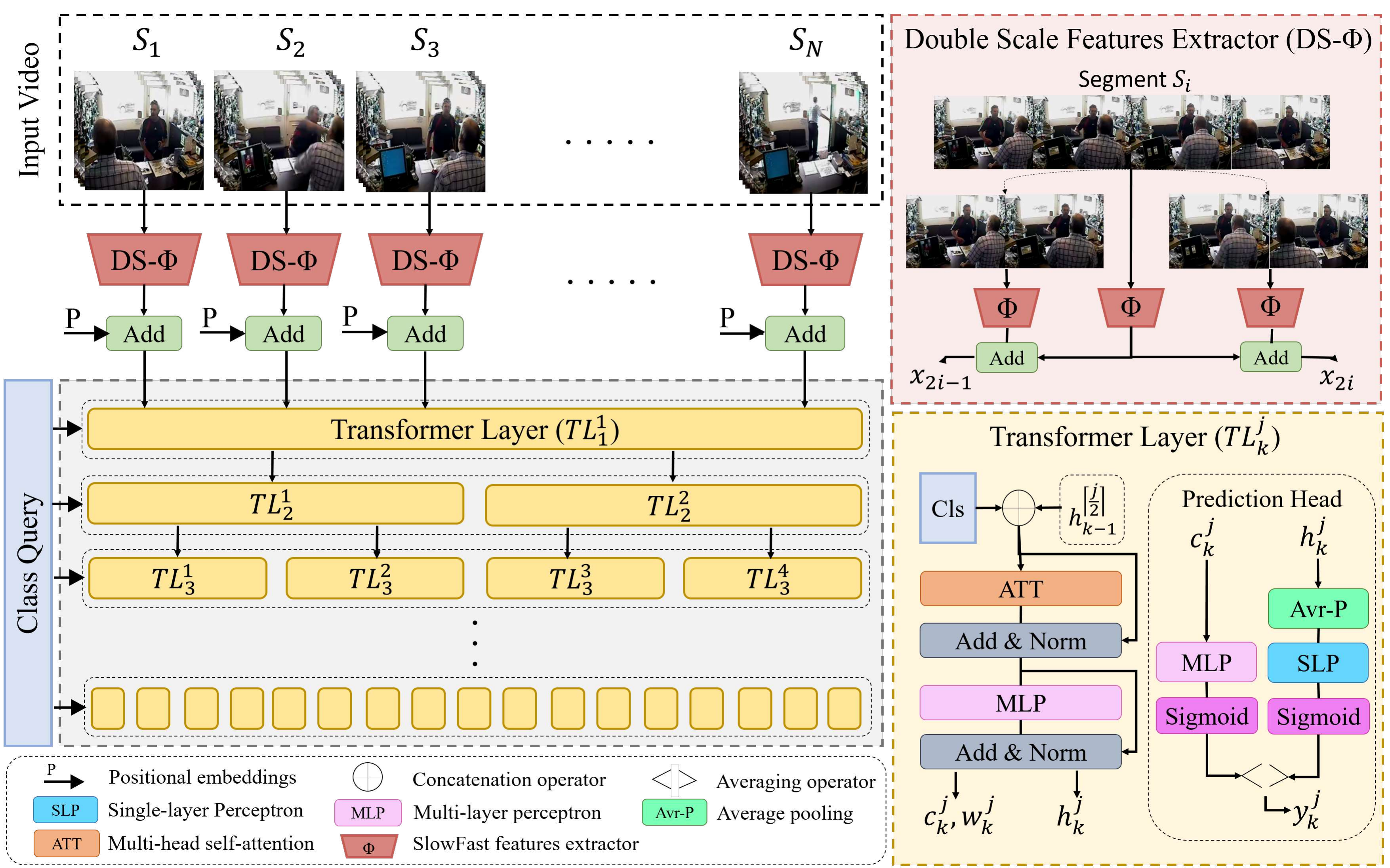}
		\caption{Our divide-and-conquer transformer-based model operates by taking the segmented video as input, where the video is divided into $N$ segments. These segments undergo feature extraction using our \textbf{Double Scale Features Extractor (DS-$\Phi$)} module. Subsequently, the extracted features are passed to the \textbf{hierarchical transformer layers} for classification. At the first level ($TL_1^1$), the model generates video classification ($y_1^1 = y_v$), and at each subsequent level ($T_k$), it produces sup-video classification $y_k^j, \quad \forall j \in [1, 2, 3, \dots 2^{k-1}]$.}
		\label{fig:anomaly_model}
	\end{figure}
	
	We consider two classification tasks: per-video classification and per-segment classification, employing our temporal divide-and-conquer model depicted in Figure \ref{fig:anomaly_model}. The model takes a video as input, which is divided into a sequence of $N$ segments. Each segment represents an action or event and consists of a set of frames. Raw images within each segment are projected into the feature space using our double-scale features extractor (DS-$\Phi$) module, followed by processing through our hierarchical transformer layers.
	
	\subsection{Double Scale Features Extractor}
	
	Our DS-$\Phi$ module is designed to enhance the temporal resolution of input segments, effectively doubling the amount of information extracted during clip processing. Given that anomaly events within a segment may exhibit varying degrees of significance across different parts of the segment, our module enables both collective and divided processing of segments. This approach allows the extraction of more focused features, leading to a deeper understanding of the underlying events. Each segment $S_i$ is split into $S_i^1$ and $S_i^2$, with the raw images of $S_i$, $S_i^1$, and $S_i^2$ projected into the feature space using $\Phi$, as in (\ref{eq:phi}), applying a feature mapping with multi-layer perceptron over the extracted features ($Feat$) by a pre-trained video-framework.
	
	\begin{equation}
		\Phi(S) = MLP(Feat(S))
		\label{eq:phi}
	\end{equation}
	
	\noindent
	Subsequently, DS-$\Phi$ allows the generation of the double temporal resolution by producing two feature vectors $\x_{2i-1}$ and $\x_{2i}$ from each segment $S_i$, as in (\ref{eq:feat_ext}), where $\x \in \mathbb{R}^{1 \times D}$ and $D$ is the features size. This approach allows for extending the temporal length to its double ($2N$).
	
	\begin{equation}
		\x_{2i-1} = \Phi(S_i) + \Phi(S_i^1) \qquad \x_{2i} = \Phi(S_i) + \Phi(S_i^2)
		\label{eq:feat_ext}
	\end{equation}
	
	\noindent
	Our model is not fixed to a specific pre-trained feature extractor; however, we consider two pre-trained models as our backbones: I3D \cite{carreira2017quo}, and Slowfast \cite{slowfast}.
	
	\subsection{Hierarchical Transformer Layers}
	
	Our divide-and-conquer approach begins with the coarse-grained task of per-video classification, iteratively breaking down the task into smaller sub-tasks (clips) until reaching the fine-grained task of per-segment classification. The features of the complete sequence $\x \in \mathbb{R}^{2N \times D}$ are combined with positional embeddings $P \in \mathbb{R}^{2N \times D}$ to retain temporal causality, forming ($\x + P$). Subsequently, this combined input is forwarded to the first level of the hierarchical transformer layers and processed through the hierarchy as shown in Figure \ref{fig:anomaly_model}.\\
	
	\noindent
	At each prediction level $k \in [1, K]$, where $K$ denotes the length of the hierarchy, the model binary-divides the received signals $h_{k-1}$ from the preceding level $k-1$ into $2^{k-1}$ splits, where we set the initial input signal $h_0$ to $\x + P$, as mentioned above. To better describe the processing flow throughout our hierarchical model, we define the following for a single split $P_k^j$ within the $k^{th}$ level in the hierarchy, where $j\in[1,2^{k-1}]$. Specifically, a split $P_k^j$ represents a cut patch from the input video that has been processed up to level $k$. This patch is then handled by a self-attention transformer layer $TL_k^j$, which processes $P_k^j$ and extracts a new hidden representation. The input to $TL_k^j$ is composed of two concatenated parts, as described in (\ref{eq:tl}), combining a fixed class query input ($Cls$), along with the hidden representation received from the preceding level $k-1$ and the parent split $P_{k-1}^{\ceil*{\frac{j}{2}}}$, defined as $h_{k-1}^{\ceil*{\frac{j}{2}}}$. The $\oplus$ in (\ref{eq:tl}) denotes the sequence concatenation operation.\\
	
	\noindent
	To this end, each transformer layer $TL_k^j$ produces three outputs for $P_k^j$, as given in (\ref{eq:ti}): $c_k^j \in \mathbb{R}^{1 \times D}$, representing the class $Cls$ encoding at the layer; $w_k^j \in \mathbb{R}^{1 \times 2N}$, denoting the attention weights inside the class query; and $h_k^j \in \mathbb{R}^{\frac{N}{2^{k-1}} \times D}$, representing the encoding vectors of the segments present in video patch $P_k^j$. The hidden encoding of the segments $h_k^j$ is next passed to the proceeding level $k+1$ for finer-grained processing of the segments, reaching up to the bottom $K^{th}$ level.
	
	\begin{equation}
		c_k^j, w_k^j, h_k^j = TL(Cls \oplus h_{k-1}^{\ceil*{\frac{j}{2}}}), \qquad \forall k \in [1, K], \qquad \forall j \in [1:2^{k-1}]
		\label{eq:tl}
	\end{equation}
	
	\subsection{Prediction Head}
	
	Finally, following each transformer layer, a prediction head generates an abnormality classification. The transformer layer at the first level $TL_1^1$ produces the per-video classification, while each subsequent layer $TL_k^j$ generates an estimation of the abnormality for the video split processed at that layer, denoted as $y_k^j$ as depicted in Figure \ref{fig:anomaly_model}. To enable the model assessing the influence of segments in predicting each sup-video (split), two prediction approaches are employed: Utilizing the generated class encoding at the layer ($c_k^j$) to produce the classification $(y_k^j)_c$, as shown in (\ref{eq:cls_pred}); and employing a sigmoid classifier head, with a single layer perceptron (SLP), on the average-pooled segments encoding vectors $h_k^j$, as illustrated in (\ref{eq:seg_pred}), generating $(y_k^j)_h$ and allowing for acquiring an activation map of the enclosed segments.
	
	\begin{equation}
		(y_k^j)_c = Sigmoid(MLP(c_k^j))
		\label{eq:cls_pred}
	\end{equation}
	
	\begin{equation}
		(y_k^j)_h = Sigmoid(SLP(AveragePooling(h_k^j)))
		\label{eq:seg_pred}
	\end{equation}
	
	\noindent
	The final prediction is computed as the average of $(y_k^j)_c$ and $(y_k^j)_h$, given by:
	
	\begin{equation}
		y_k^j = Average((y_k^j)_c + (y_k^j)_h)
	\end{equation}
	
	\subsection{Localization Approach}
	
\begin{figure}[t]
		\centering
		\includegraphics[width=\linewidth]{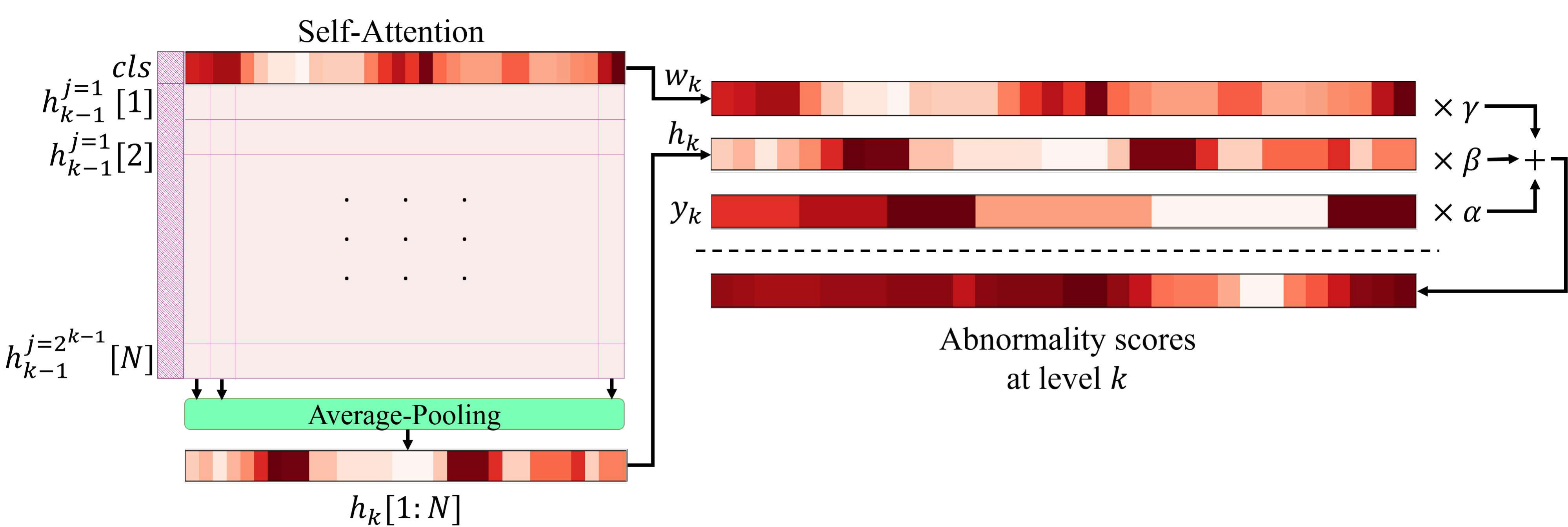}
		\caption{\textbf{Visualization of our localization Approach.} Assuming the localization is conducted at level $k$, $w_k$ is obtained from the attention weights of the class query inside our self-attention layers, $h_k$ is the averaged-pooled encodings produced by the transformer, and $y_k$ is the stacked sub-predictions at $k$. The estimated abnormality is computed as in Equation (\ref{eq:seg_cls}).}
		\label{fig:localization_technique}
	\end{figure}

	The model breaks the video prediction into a set of sub-predictions, where we aim to measure the influence of a segment $S_i$ in all of its corresponding sub-predictions. To capture such influence, we rely on three measuring factors, as illustrated in Figure \ref{fig:localization_technique}:
	
	\begin{enumerate}
		\item The abnormality prediction in the corresponding patches across the different levels ($p_i$); the probability of $S_i$ to be an anomaly segment is monotonically increasing with the corresponding probability of a parent clip. Therefore, this probability is measured as the averaged predictions of the parent clips across the $K$ levels, given by:
		
		\begin{equation}
			p_i = \underset{k \in [1, K]}{Average} \quad y_k^j, \qquad j = \lceil \frac{i}{2^{k-1}}\rceil
			\label{eq:pi}
		\end{equation}     
		
		\item The activation effect of $S_i$ in the prediction of parent patches ($a_i$); The higher the activation of the segment in an anomaly class, the higher its probability of being an anomaly. The averaged activation across the levels is given by:
		
		\begin{equation}
			a_i = \underset{k \in [1, K]}{Average} \quad h_k[i]
			\label{eq:ai}
		\end{equation}
		
		Where $h_k$ is the stacked, average-pooled hidden representations of the $2^{k-1}$ transformer layers at level $k$, and $h_k[i]$ is the corresponding representation of $S_i$. 
		\vspace{0.5cm}
		
		\item The corresponding attention weights of the segment $S_i$ in the class query ($t_i$); Again, the higher attention given to a segment during the anomaly prediction $(y_k^j)_c$, the greater its influence and the higher its probability of being an anomaly. Therefore the attention maps of the class query in the corresponding splits are averaged across the levels to yield the attentive weight of $S_i$ in the prediction:
		
		\begin{equation}
			t_i = \underset{k \in [1, K]}{Average} \quad w_k[i]
			\label{eq:ti}
		\end{equation}
		
		Where, $w_k$ is the concatenated attention weights across the employed attentions at level $k$, and $w_k[i]$ is the weight of the corresponding segment $S_i$. 
	\end{enumerate}
	
	\noindent
	Based on these three factors, $p_i$, $a_i$, and $t_i$, an aggregation of the abnormality estimation score $e_i$ of $S_i$ is calculated as the weighted average of the normalized factors, as specified in (\ref{eq:seg_cls}). Here, $\alpha$, $\beta$, and $\gamma$ represent the weighting parameters. Finally, considering the tendency of anomaly segments to occur in clusters within the video, we smooth the estimated probabilities across the video using a moving average. Subsequently, we apply spike filtering to eliminate potential outlier scores.
	
	\begin{equation}
		e_i = \alpha p_i + \beta a_i + \gamma t_i
		\label{eq:seg_cls}
	\end{equation}
	
	\section{Experimental Results}
	
	\subsection{Dataset}
	
	We evaluate our model on two datasets: UCF-Crime dataset \cite{dataset_anomaly} and ShanghaiTech \cite{liu2018future}. UCF-Crime dataset comprises 1,900 surveillance videos spanning over 128 hours, with 13 distinct anomaly behaviors such as abuse, assault, accident, fighting, robbery, etc. On the other hand, ShanghaiTech is a smaller dataset, comprising 437 videos, with 130 anomaly videos and 307 normal videos. The videos are partitioned into 32 segments each, with ground truth per-segment class labels provided exclusively for the segments in the test split of the data.
	
	\subsection{Evaluation Metrics}
	
	Following the established evaluation protocol for this task \cite{ss_anomaly1}, we utilize the Area under the ROC Curve (AUC-ROC) metric to assess the localization performance of anomaly segments. Additionally, we report accuracy and F1-score metrics to measure the model's ability to distinguish between normal and abnormal segments.
	
	\subsection{Implementation details}
	For both of our evaluation datasets, we set the number of segments $N$ to 32 segments. The extracted feature sizes of the pre-trained video frameworks are $2048$ and $2304$ for I3D and SlowFast, respectively, while the mapped features size $D$ is set to $256$ and $288$. Each transformer layer includes $4$ self-attention heads, and the hierarchical model consists of $K=6$ levels. The Multi-Layer Perceptron (MLP) module inside the prediction head consists of three layers with sizes $128$, $64$, and $32$. The dropout rate for all model layers is set to $0.1$. The model is trained for 100 epochs using a Stochastic Gradient Descent (SGD) optimizer, with a learning rate of $0.01$, and binary cross-entropy loss function. During training, all children split within the hierarchical model inherit their anomaly class label from the parent video. During inference, our localization approach is applied using the weighting parameters $\alpha$, $\beta$, and $\gamma$, configured to $0.9$, $0.05$, and $0.05$, for UCF-crime. While $0.65$, $0.3$, and $0.05$ are used for ShanghaiTech.
	
	\begin{table}[t]
		\centering
		\caption{\textbf{Comparison with the state-of-the-art works on UCF-Crime}. For a fair comparison, we separate the performance based on the application of pseudo-labeling (PL). Our model uniquely overlooks both MIL and pseudo-labeling techniques. Instead, it employs a hierarchical measuring technique for abnormality weights of the per-video segments, outperforming all previous works w/o PL. However, the best performances are in favor of the utilization of tailored pseudo labels for the dataset, with a comparable performance of our HCAM-former with most PL-works.} \vspace{-0.3cm}
		\begin{adjustbox}{width=\linewidth,center}
			\setlength{\tabcolsep}{3pt}
			\begin{tabular}{l c c c c c c c c}
				\toprule
				& \multirow{2}{*}{MIL} & Segments && Features & \multirow{2}{*}{Classifier} && \multicolumn{2}{c}{AUC}\\
				
				\cline{8-9}
				
				&& Activation && Extractor &  && w/o PL & PL\\
				\midrule
				MIL (2018) \cite{dataset_anomaly} & \cmark & \xmark && C3D & FC && 75.41 & -\\
				
				TCN-IBL (2019) \cite{ss_anomaly2} & \cmark & \xmark && C3D & TCN && 78.66 & -\\
				
				MIL-MA (2019) \cite{ss_anomaly1} & \cmark & \xmark && PWCNet & TAN && 79.00 & -\\ 
				
				
				GC-LNC (2019) \cite{ss_anomaly5} & \xmark & \xmark && TSN & GCN && 70.87 & 82.12\\
				
				CPL (2020) \cite{cam_anomly} & \cmark & \cmark && BN-Inception & ResNet+KNN && - & 79.31\\
				
				MIL-MIST (2021) \cite{ss_anomaly3} & \cmark & \xmark && I3D & Attention && 73.33 & 82.30\\
				
				MIL-AUG (2022) \cite{ss_anomaly4} & \cmark & \xmark && SlowFast & FC && 79.37 & 81.24\\
				
				RTFM (2022) \cite{tian2021weakly} & \cmark & \cmark && I3D & PDC && - & 84.30\\
				
				MSL (2022) \cite{li2022self} & \cmark & \xmark && VideoSwin & Transformer && - & 85.62\\
				
				CU-Net (2023) \cite{zhang2023exploiting} & \cmark & \xmark && I3D & FC && - &\textbf{\textcolor{red}{86.22}}\\
				
				C2FPL (2024) \cite{al2024coarse} & \cmark & \xmark && I3D & FC && 72.70 & 85.50\\
				
				\midrule
				
				\multirow{2}{*}{HCAM-former (Ours)} & \multirow{2}{*}{\xmark} & \multirow{2}{*}{\cmark} && I3D & \multirow{2}{*}{H-Transformer} && \textbf{\textcolor{blue}{78.30}} & - \\
				& & && SlowFast & && \textbf{\textcolor{blue}{79.47}} & - \\
				
				\bottomrule
				
			\end{tabular}
		\end{adjustbox}
		\label{tab:anomaly_comarison}
	\end{table}
	
	\begin{table}[t]
		\centering
		\caption{\textbf{Comparison with the state-of-the-art works on ShanghaiTech}. Again, the HCAM-former outperforms all w/o PL models while achieving comparable performance to the best PL works.} \vspace{-0.3cm}
		\begin{adjustbox}{width=\linewidth,center}
			\setlength{\tabcolsep}{3pt}
			\begin{tabular}{l c c c c c c c c}
				\toprule
				& \multirow{2}{*}{MIL} & Segments && Features & \multirow{2}{*}{Classifier} && \multicolumn{2}{c}{AUC}\\
				
				\cline{8-9}
				
				&& Activation && Extractor &  && w/o PL & PL\\
				\midrule
				MIL (2018) \cite{dataset_anomaly, tian2021weakly} & \cmark & \xmark && C3D & FC && 85.33 & -\\
				
				GC-LNC (2019) \cite{ss_anomaly5} & \xmark & \xmark && TSN & GCN && 80.83 & 84.44\\
				
				AR-Net (2019) \cite{wan2020weakly} & \cmark & \xmark && I3D & FC && 91.24 & - \\
				
				MIL-MIST (2021) \cite{ss_anomaly3} & \cmark & \xmark && I3D & Attention && - & 94.38\\
				
				RTFM (2022) \cite{tian2021weakly} & \cmark & \cmark && I3D & PDC && - & 97.21\\
				
				MSL (2022) \cite{li2022self} & \cmark & \xmark && VideoSwin & Transformer && - & \textbf{\textcolor{red}{97.32}}\\
				
				\midrule
				
				HCAM-former (Ours) & \xmark & \cmark && I3D & H-Transformer && \textbf{\textcolor{blue}{93.29}} & - \\
				
				\bottomrule
				
			\end{tabular}
		\end{adjustbox}
		\label{tab:anomaly_comarison_sh}
	\end{table}
	
	\subsection{Results}
	
	We conduct a comparative analysis of our model against state-of-the-art works in anomaly detection. Table \ref{tab:anomaly_comarison} presents a comparison on the UCF-Crime dataset, highlighting the structural differences between models to ensure a fair evaluation. This includes whether the learning process employs multiple-instance learning, benefits from segment activation, the feature extractor used, and the final classification module of each detection methodology. Additionally, we differentiate the performance of the models based on the application of pseudo-labeling. As shown in the table, our HCAM-former outperforms all other techniques when pseudo-labeling is excluded, underscoring the efficacy of our method in interpreting observed events and distinguishing anomalies. However, while our model achieves promising results, it generally falls behind the performance of pseudo-labeling approaches. These methods handle weakly supervised data by generating estimated labels for refinement and use as ground truth, and such iterative learning techniques enhance anomaly pattern recognition. As mentioned earlier, our aim is to provide a more generic solution that is not as dataset-specific as the pseudo-labeling technique. It is noteworthy that our model still outperforms the CPL \cite{cam_anomly} approach, which utilizes segments CAM (only hidden representation CAM $a_i$ in our approach), in addition to employing both MIL and pseudo-labeling techniques. Similarly, Table \ref{tab:anomaly_comarison_sh} reports the results on the ShanghaiTech dataset. Again, our model achieves state-of-the-art performance when excluding pseudo-labels while reaching a compatible performance with the PL approaches and even outperforms the GC-LNC(PL) technique.\\
	
	\noindent
	The provided comparison results prove the effectiveness of our proposed approach in identifying anomaly events from normal ones solely based on the temporal progression of events within the parent context. Although it has a degraded performance compared to the pseudo-labeling models, our model provides a more generic end-to-end solution that overlooks tailoring pseudo-labels for the evaluation datasets, which is a promising direction with respect to real-world applications.
	
				
				
				
				
				
				
				
				
				
				
	\subsection{Ablation Study}
	
	We conducted an ablation study to assess the impact of different components of our proposed model on per-video classification performance on UCF-Crime, as summarized in Table \ref{tab:per-video_ablation}. Since the model's ability to recognize videos with anomaly events from normal videos depends on its capability to extract anomaly behaviors within the videos, achieving high per-video performance is crucial to ensure the accuracy of the estimated segment scores by the model. This ablation study examines three main components of our model: 1) The hierarchical structure (comparing $K=1$ against $K=6$), where $K=1$ denotes that only the first level of HCAM-former is trained and used for inference depending on $a_i$ and $t_i$. While $K=6$ denotes the utilization of the whole hierarchy. 2) The inclusion of the class query in the transformer layers. 3) The double-scale features extractor module DS-$\Phi$.\\
	
	\noindent
	The hierarchical structure has the most significant impact, contributing to an approximately $3\%$ increase in the F1-score  and $9\%$ in AUC, compared to the model without the hierarchy. Additionally, the class query and the DS-$\Phi$ module notably improve the model's performance. As a result, the highest-performing configuration is achieved by including all components of our model.\\
	
	\begin{table}[t]
		\centering
		\caption{\textbf{Ablation on per-video classification performance (UCF-Crime).} When $K=1$, the model operates without the hierarchical structure. In this configuration, only the first transformer layer of our model is utilized to make predictions and measure the abnormality scores of the segments.}
		\begin{adjustbox}{width=\linewidth,center}
			\setlength{\tabcolsep}{8pt}
			\begin{tabular}{@{\hspace{1.5em}} c c c c c c c c c}
				\toprule
				\multicolumn{2}{c}{Hierarchical Transformer}&& Class & \multirow{2}{*}{DS-$\Phi$} && \multirow{2}{*}{ACC} & \multirow{2}{*}{AUC} & \multirow{2}{*}{F1}\\
				\cline{1-2}
				$K=1$ & $K=6$ && Query & &&&&\\
				\midrule
				
				\cmark & \xmark && \xmark & \xmark && 83.97 & 83.89 & 83.09\\
				
				\cmark & \xmark && \cmark & \xmark && 85.02 & 84.83 & 83.40\\
				
				\cmark & \xmark && \cmark & \cmark && 85.02 & 84.90 & 83.89\\[0.5ex]
				
				\cdashline{1-9}\noalign{\vskip 0.5ex}
				
				\xmark & \cmark && \cmark & \cmark && \textbf{87.12} & \textbf{92.44} & \textbf{86.74}\\
				\bottomrule 
			\end{tabular}
		\end{adjustbox}
		\label{tab:per-video_ablation}
	\end{table}
	
	\begin{table}[t]
		\centering
		\caption{\textbf{Ablation on the localization performance (UCF-Crime).} Where, $a_i$ denotes the class activation maps in (\ref{eq:ai}), $t_i$ is the attention weights of the class query in (\ref{eq:ti}), and $p_i$ is probability estimation in (\ref{eq:pi}). \vspace{-0.5cm}}
		\begin{adjustbox}{width=\linewidth,center}
			\setlength{\tabcolsep}{9pt}
			\begin{tabular}{@{\hspace{1.5em}}c c c c c c c c c c}
				\toprule
				\multicolumn{2}{c}{Hierarchical Transformer}&& \multirow{2}{*}{$a_i$} & \multirow{2}{*}{$t_i$} & \multirow{2}{*}{$p_i$} && \multirow{2}{*}{ACC} & \multirow{2}{*}{AUC} & \multirow{2}{*}{F1}\\
				\cline{1-2}
				$K=1$ & $K=6$ && & &&&&&\\
				\midrule
				
				\cmark & \xmark && \cmark & \xmark & \xmark && \textbf{73.73} & 72.83 & 40.11\\
				
				\cmark & \xmark && \xmark & \cmark & \xmark && 70.09 & 72.28 & 38.17\\
				
				\cmark & \xmark && \cmark & \cmark & \xmark && 73.18 & 73.87 & \textbf{40.65}\\[0.5ex]
				
				\cdashline{1-10}\noalign{\vskip 0.5ex}
				
				\xmark & \cmark && \cmark & \cmark & \cmark && 66.54 & \textbf{79.47} & 37.78\\
				\bottomrule 
			\end{tabular}
		\end{adjustbox}
		\label{tab:per-segment_ablation}
	\end{table}
	
	\noindent
	In Table \ref{tab:per-segment_ablation}, we ablate our localization technique. By excluding hierarchical predictions, we compare the localization performance using the class activation of the first transformer layer in the hierarchy ($a_i$) and the attention weights of the class query of the same layer, both individually and integrated. The results in the table indicate that CAM estimation slightly outperforms attention weights in terms of localization. However, combining both estimations yields even better localization performance. Integrating hierarchical predictions as abnormality probability estimation achieves the highest localization performance, with an improvement of approximately $5\%$ in AUC. As can be noticed, when the AUC metric is increased, accuracy and F1 score suffer a bit of degradation. This could be explained by the severe imbalance in the number of anomaly segments compared to normal ones. Therefore, maximizing the number of correctly localized anomalies leads to an increase in the number of misclassified normals. However, correctly localizing anomalies presents a higher priority, and therefore, the AUC metric is more relevant for the task.\\
	
	\noindent
	Finally, Table \ref{tab:feat_ablation} evaluates the impact of the feature extractor on anomaly localization performance. The results favor the SlowFast feature extractor over the I3D features, even when the number of heads in our transformer layers is increased to 8 for the I3D features. This finding demonstrates the superior capability of the SlowFast framework to interpret spatial and temporal information in the input images, enabling our model to better understand and localize the anomaly events.
	
	\begin{table}[t]
		\centering
		\caption{\textbf{Ablation on the features extractor (UCF-Crime).} The ACC, AUC, and F1 are reported for both per-video classification and per-segment classification.}
		\begin{adjustbox}{width=\linewidth,center}
			\setlength{\tabcolsep}{10pt}
			\begin{tabular}{c c c c c c c c c}
				\toprule
				\multirow{2}{*}{Features}&& \multicolumn{3}{c}{Per-Video} && \multicolumn{3}{c}{Per-Segment}\\
				\cline{3-5} \cline{7-9}
				&& ACC & AUC & F1 && ACC & AUC & F1\\
				
				\midrule
				
				I3D && 80.21 & 77.6 & 82.9 && 65.27 &78.30 & 23.70\\
				SlowFast && 87.12 & 92.44 & 86.74 && 66.54 & 79.47 & 37.78\\
				
				\bottomrule
			\end{tabular}
		\end{adjustbox}
		\label{tab:feat_ablation}
	\end{table}
	
	\vspace{-0.4cm}
	\subsection{Qualitative Examples}
	\vspace{-0.1cm}
	
	\begin{figure}[t]
		\centering
		\includegraphics[width=\linewidth]{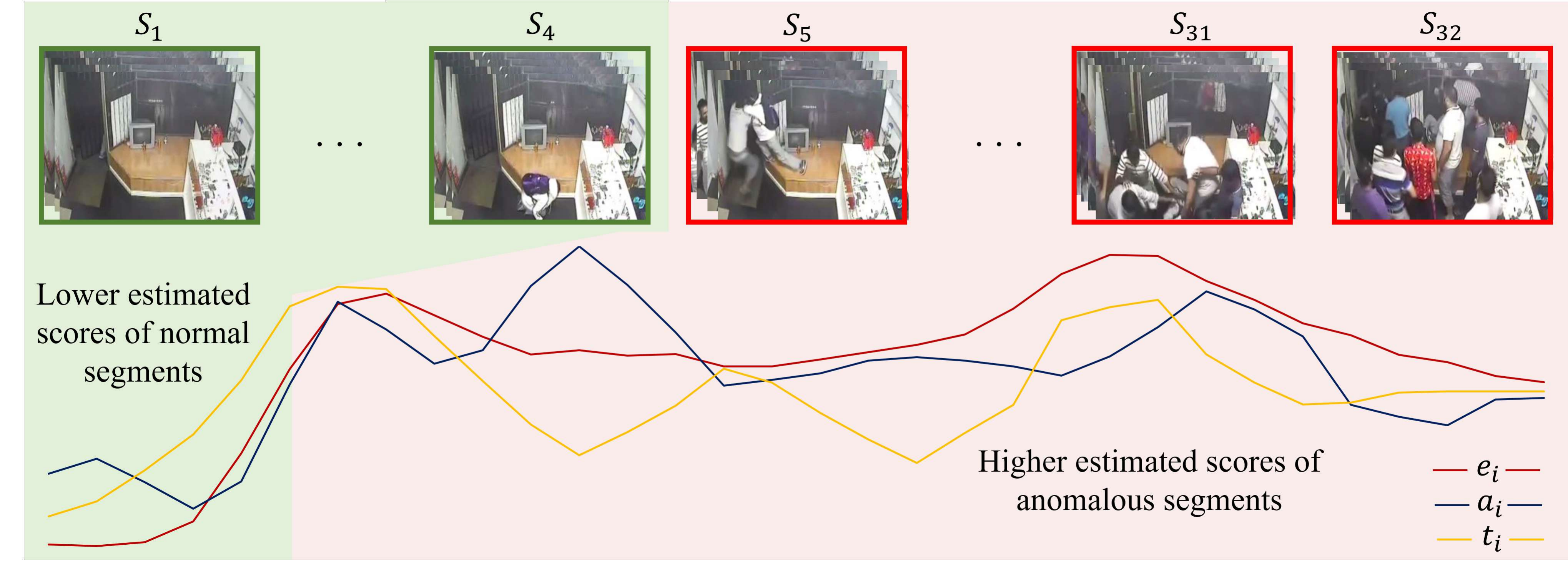}
		\caption{\textbf{Qualitative example on accurate anomaly localization from UCF-Crime.} Anomaly score estimations are provided for the anomaly action \textit{"Assault"}, which begins at segment $S_5$ and continues until the final segment $S_{32}$. Here, $e_i$ represents the score estimated in (\ref{eq:seg_cls}), $a_i$ denotes the segment's activation as defined in (\ref{eq:ai}), and $t_i$ refers to the attention weights described in (\ref{eq:ti}). \vspace{-.2cm}}
		\label{fig:qualitative_ex_anomaly}
	\end{figure}
	
	Figure \ref{fig:qualitative_ex_anomaly} illustrates the estimated anomaly scores for segments within an anomaly video featuring an \textit{"Assault"} event. The video begins with normal events during the first segments $[S_1-S_4]$; then the anomalous event starts at $S_5$. The plotted anomaly scores for all three abnormality measuring factors exhibit lower values at the beginning, gradually increasing as the anomaly action unfolds. The aggregated anomaly estimation $e_i$ yields smoother estimations across the segments, benefiting from the fusion of the used factors. While both the activation map $a_i$ and attention weights $t_i$ show slightly decreased scores during the anomaly event, they remain higher than those of the normal segments. Towards the end of the video, as the \textit{"Assault"} event subsides, the anomaly estimation begins to decrease. This demonstrates the model's capability to interpret events depicted in the segments and distinguish anomalies, relying solely on the information gained during video classification.\\
	
\begin{figure}[t]
		\centering
		\includegraphics[width=\linewidth]{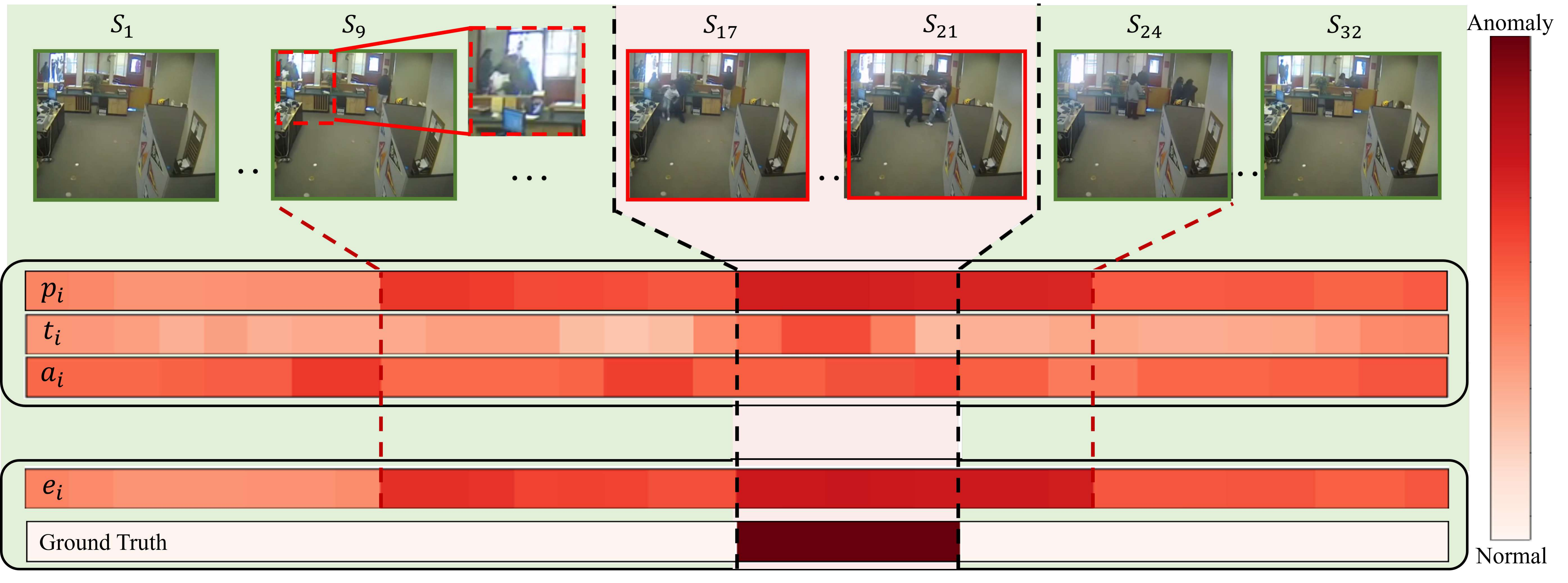}
		\caption{\textbf{Qualitative example on slightly misled anomaly localization from UCF-Crime.} The heat maps illustrate anomaly score estimations for the \textit{"Arrest"} event, which initiates at segment $S_{17}$ and extends through segment $S_{21}$. The illustrated heat maps are $p_i$, $t_i$, $a_i$, the aggregated estimation $e_i$, and the ground truth labeling of the segments.}
		\label{fig:qualitative_ex_anomaly2}
	\end{figure}

	\noindent
	The qualitative example depicted in Figure \ref{fig:qualitative_ex_anomaly2} aims to illustrate instances where the model may misinterpret anomaly segments. To provide a more sensitive analysis, heat maps are employed for visualization, where an \textit{"Arrest"} event is occurring from segment $S_{17}$ to $S_{21}$. Initially, all abnormality measures exhibit relatively lower scores during the video's early segments, indicating normalcy. However, starting from segment $S_9$, scores begin to rise, particularly for $p_i$ and $a_i$. Notably, an event at $S_9$ depicts a violent interaction in the left corner of the frame, leading to elevated abnormality scores for segments $[S_9 - S_{16}]$. Then, as the actual \textit{"Arrest"} event unfolds, the model accurately assigns the highest abnormality scores; however, the highest scores are persisted until segment $S_{24}$. This is due to another misleading event featuring multiple individuals gathered around the arrest location, contributing to the sustained high scores beyond the actual event. It is worth noting that throughout the observation, attention scores provided by $t_i$ closely align with the ground truth, yet they tend to emphasize only the most significant segments, potentially overlooking some anomalous events.\\
	
	\noindent
	From both examples, it is evident that the factor with the most effective capability in recognizing anomaly events is the estimated probability using our hierarchical predictions.
	
	\vspace{-0.25cm}
	\section{Conclusion}
	
	In this work, we tackle the challenge of anomaly detection in weakly-supervised datasets. We redefine the anomaly detection task from per-segment to per-video classification, leveraging the temporal progression of videos to localize anomaly events effectively. Our proposed temporal divide-and-conquer hierarchical transformer model surpasses state-of-the-art non-pseudo-labeling methods and achieves promising results compared to tailored pseudo-labeling approaches. These findings demonstrate the promising capability of our model to process temporal video contexts, comprehend observed events, and correctly localize anomalies. Although the current performance of our proposed approach does not yet match the higher performance achieved by tailoring pseudo-labels for the datasets, we explore a more explainable and general solution of video-level anomaly localization, showcasing the promising capability of such a technique in interpreting and localizing abnormal events. Therefore, our future work will focus on enhancing video-level anomaly localization, allowing for a more in-depth study of its performance, explainability, and generality compared to pseudo-labeling approaches.
	
	%
	%
	%
	\bibliographystyle{splncs04}
	\bibliography{main}
\end{document}